\documentclass[preprint,nopreprintline]{elsarticle}

\usepackage[a4paper,textwidth=440pt,textheight=640pt,centering,headheight=14pt]{geometry}

\usepackage[T1,OT1]{fontenc}  
\usepackage{lmodern}  
\usepackage[scaled=0.90]{beramono}

\newcommand{\bera}{\fontencoding{T1}\fontfamily{fvm}\selectfont}
\usepackage{amssymb}
\usepackage{amsmath}
\usepackage{amsfonts}
\usepackage{bm}
\usepackage{graphicx}
\graphicspath{{./}}
\usepackage{booktabs}
\usepackage{array}
\usepackage{xcolor}

\definecolor{revisionred}{RGB}{190,0,0}

\newif\ifdarkmode
\darkmodefalse  
\ifdarkmode
  \pagecolor{black!90}
  \AtBeginDocument{%
    \color{white!90}%
    \captionsetup{font={color=white!90},labelfont={color=white!90,bf}}%
  }
\fi

\usepackage{mathtools}
\usepackage{caption}
\usepackage{placeins}
\usepackage{url}
\usepackage{microtype}
\usepackage{multirow}

\usepackage[most]{tcolorbox}
\newtcolorbox{promptbox}[1][]{enhanced,breakable,
  colback=black!2,colframe=black!45,boxrule=0.4pt,arc=1.5pt,
  left=5pt,right=5pt,top=2.5pt,bottom=2.5pt,boxsep=1pt,
  before skip=4pt,after skip=4pt,
  fonttitle=\footnotesize\bfseries,coltitle=black,
  attach title to upper={\par\smallskip},#1}

\usepackage[left]{lineno}
\nolinenumbers
\journal{}

\newif\ifanonymous
\anonymousfalse

\renewcommand\appendixname{Appendix}

\usepackage[hidelinks]{hyperref}

\begin{document}

\begin{frontmatter}

\title{Geological text descriptions in ill-posed inverse problems: insights from learned hydraulic-conductivity inversion}

\ifanonymous\else
\author[label1]{Taiga Saito\corref{cor1}}
\ead{taiga.saito.r3@dc.tohoku.ac.jp}
\cortext[cor1]{Corresponding author}

\author[label1]{Yu Otake}
\author[label1]{Daijiro Mizutani}
\author[label1]{Sopheakpolin Mom}

\affiliation[label1]{
    organization={Department of Civil and Environmental Engineering, Tohoku University},
    addressline={6-6-06 Aramaki Aoba, Aoba-ku},
    city={Sendai},
    state={Miyagi},
    postcode={980-8579},
    country={Japan}
}
\fi

\begin{abstract}
Hydraulic-conductivity inversion is ill posed: even complete head observations can leave structural ambiguity.
Geological text descriptions can supply additional information about subsurface structure to constrain reconstruction.
However, it remains unclear when descriptions improve reconstruction and how solvers use their stated content.
We examine these questions in learned inversion using a synthetic Darcy-flow dataset with idealised descriptions, varying observation density and flow direction.
After sparse-observation training, descriptions can improve reconstruction over models trained without them when observations leave structural ambiguity, though gains vary across conditions and runs.
Instance-specific content can improve reconstruction beyond field-type information.
Editing this content shifts reconstructions toward the stated geometry of selected features, less strongly as observations become more informative about those features.
We do not establish a reconstruction advantage of text over numerical inputs encoding the same geological information.
Finally, we discuss how this complementarity could guide the joint design of measurements and geological knowledge acquisition in site investigations.
\end{abstract}


\end{frontmatter}

\section{Introduction}
\label{sec:intro}

Ill-posed inverse problems arise when physical observations do not uniquely or stably determine the quantities to be inferred \cite{tarantola2005,aster2018}.
Hydraulic-conductivity inversion is one such problem: spatially varying conductivity is inferred from hydraulic-head observations using a Darcy-flow model, yet different fields can remain compatible with the available heads \cite{zhou2014geostatistical}.
To address this ill-posedness, regularised inversion supplements data fitting with constraints such as spatial smoothness \cite{aster2018}.
Geological knowledge can inform these constraints \cite{linde2015geological}, for example through geostatistical models and representations learned from training images \cite{laloy2018training,li2015localglobal}.
Markov random fields have also been used in site characterisation to incorporate spatial continuity and geological prior knowledge into stratigraphic models built from borehole records \cite{tian2026gmm}.
Alongside such prior information, measurements of different types can be combined; for example, machine learning has been used to integrate seismic reflection data with sparse cone penetration test data in quantitative ground modelling \cite{qu2026ground}.
In direct supervised inversion, learned solvers are trained on simulated pairs of conductivity fields and physical responses to map observations to conductivity fields \cite{triplett2025deep,zhu2026multimodal}; the resulting mappings therefore reflect the geological patterns represented in the training data.
To supplement this knowledge of geological patterns, information specific to the field being reconstructed can be supplied as conditioning inputs alongside its head observations.

Geological text descriptions offer one such conditioning input.
In geotechnical site characterisation, site-specific measurements are often sparse, incomplete, and uncertain \cite{phoon2022challenges}.
This motivates making fuller use of site-specific geological information documented in borehole logs, site investigation reports, and expert interpretations.
Large language models are being explored as ways to retrieve, organise, and use such textual information alongside quantitative data \cite{wu2024futureproofing,wu2025agents,qian2025llm}.
Retrieval-augmented language models have been used to standardise formation types in heterogeneous borehole records \cite{yan2026standardization}.
Large language models have also been prompted with a small set of worked examples of geological reasoning to construct geological cross-sections from sparse site investigation data \cite{li2025fewshot}.
For inversion, a description makes the supplied geological assumptions explicit and open to human review, and can provide additional constraints when it contains information beyond the physical observations.
Semantic regularisation provides a precedent for using descriptions as prior information in inverse problems: in electromagnetic inversion, knowledge expressed in language has been incorporated alongside physical measurements to guide reconstruction \cite{zhang2024semantic,chen2024semanticem}.
To incorporate such descriptions into learned conductivity inversion, sentence encoders can represent them as numerical vectors that serve as conditioning inputs \cite{reimers2019sentencebert}.
Such conditioning combines two forms of information: descriptions state interpretations of subsurface structure, whereas physical observations can be compared with responses predicted by an assumed forward model.
How well physical observations can test a stated interpretation therefore depends on the sensitivity of the predicted response to the described features.

The use of geological descriptions as conditioning inputs raises two questions for learned conductivity inversion: under which observation conditions do they improve reconstruction, and to what extent does a learned solver use their stated content?
For the first question, the structural ambiguity left by physical observations depends on their coverage and on the sensitivity of the measured response to geological features \cite{guillaume2019identifiability}.
Additional measurements at informative locations can reduce ambiguity due to limited coverage \cite{zhao2019sampling}.
When distinct structures produce the same physical response under given forcing and boundary conditions, however, the inverse problem remains non-unique and therefore ill posed even with complete observations of that response.
Distinguishing such structures requires additional information, for example from another physical response, observations under different forcing or boundary conditions, or geological descriptions.
For the second question, a description may convey both a broad structural type and information specific to the field being reconstructed.
Joint modelling of soil-class labels and continuous soil parameters has been used to characterise classification-dependent relationships among soil properties \cite{saito2025midm}.
An improvement over a model without descriptions may arise from identifying the structural type alone and therefore does not establish which instance-specific details, if any, the solver uses.
Analyses of how individual inputs affect predictions can thus complement performance evaluation, as illustrated for machine-learning predictions in hydrometeorology \cite{huang2023beyond}.
A gain over a model without descriptions also leaves open whether natural language has an advantage over a numerical representation of the same information.

To examine these questions, this study introduces a synthetic Darcy-flow dataset with six types of conductivity fields and idealised descriptions generated by a language model from selected field-generation attributes.
Because the conductivity fields are known, reconstruction can be evaluated against ground truth while observation conditions and stated content are controlled; the simulated heads carry no added measurement noise.
We vary observation density, which changes the coverage of the head field, and flow direction, which can change the sensitivity of the head to structural features.
For the first question, we compare a solver conditioned on sentence embeddings from a frozen encoder with a matched solver retrained without descriptions under these observation conditions.
For the second question, we compare the sentence-conditioned solver with a one-hot control encoding only the field type to assess the reconstruction value of instance-specific content beyond field-type information.
The sentence-conditioned solver is also compared with a Numeric facts model supplied with the same stated attributes to assess reconstruction under sentence and numerical conditioning.
For these model comparisons, we evaluate differences in reconstruction error by field type, with results paired by training seed.
To complement this performance evaluation, we use ridge probes to assess the linear accessibility of selected attributes in sentence embeddings and controlled text edits to assess the influence of those attributes on reconstruction.
Finally, we discuss how geological text descriptions could be used in other ill-posed inverse problems, considering the information they add, how solvers use them, and their reliability.
The dataset, fixed splits, controlled text edits, and data-generation and loading code are available on Zenodo \cite{darcy_dataset}.

\section{Problem Formulation and Synthetic Dataset}
\label{sec:methods_data}

We formulate the recovery of a spatially varying hydraulic conductivity field $K$ from observations of hydraulic head $h$, together with a natural-language geological description $t$, as a supervised learning task. The dataset is fully synthetic: head fields are computed from the Darcy equation, and descriptions verbalise selected attributes derived from the field-generation parameters rather than from the realised field.

\subsection{Language-Conditioned Inverse Problem}
\label{sec:setting_problem}

Under steady, source-free Darcy flow, the conductivity field $K$ and hydraulic head $h$ satisfy
\begin{equation}\label{eq:darcy}
  -\nabla \cdot (K \nabla h) = 0
\end{equation}
on the unit square $\Omega = [0,1]^2$, with spatial coordinates in metres.
The domain represents a two-dimensional geological cross-section, with $x$ horizontal and $y$ increasing upward.
The left and right boundaries are held at $h=1$ and $h=0$, respectively, with zero normal flux through the upper and lower boundaries.
We refer to this configuration as horizontal flow, parallel to the strata of the Layered class (\S\ref{sec:setting_fields}).
Under these boundary conditions, the horizontally stratified Layered fields form an idealised extreme case: the exact solution is $h=1-x$, independent of the number, positions, and thicknesses of the strata.
To make the head sensitive to stratum position, we also consider vertical flow, perpendicular to the strata: $h=1$ at the bottom, $h=0$ at the top, and zero normal flux through the left and right boundaries.
In both configurations, $h$ is expressed relative to the low-head boundary and scaled by the imposed head difference.

Recovering $K$ from hydraulic head alone is ill posed: distinct conductivity fields can produce identical or nearly identical head fields, even when the complete head field is available.
For example, under either flow configuration, multiplying $K$ by any positive constant leaves $h$ unchanged.
Sparse observations can introduce additional ambiguity in recovering $K$.
To examine when geological descriptions improve conductivity reconstruction, we conduct controlled synthetic experiments that vary observation density, flow direction, and description content.
We prescribe the clay and sand conductivities and the clipping bounds of the Continuous field (\S\ref{sec:setting_fields}).

Let $h_{\mathrm{obs}}$ denote the input constructed from the available hydraulic-head observations, and let $t$ denote the corresponding geological description.
The language-conditioned solver returns the point estimate
\begin{equation}
  \hat{K} = G_\phi(h_{\mathrm{obs}}, t),
\end{equation}
where $\phi$ denotes the learned model parameters.
Both $h_{\mathrm{obs}}$ and $\hat{K}$ are represented as $64 \times 64$ arrays, with the input construction and conductivity normalisation described in \S\ref{sec:methods_arch}.
We assess the value of $t$ by comparing this solver with a matched baseline trained without descriptions (\S\ref{sec:methods_arms}).

\subsection{Geologically Motivated Fields and Darcy-Flow Simulation}
\label{sec:setting_fields}

\begin{figure}[t]
  \centering
  \includegraphics[width=\textwidth]{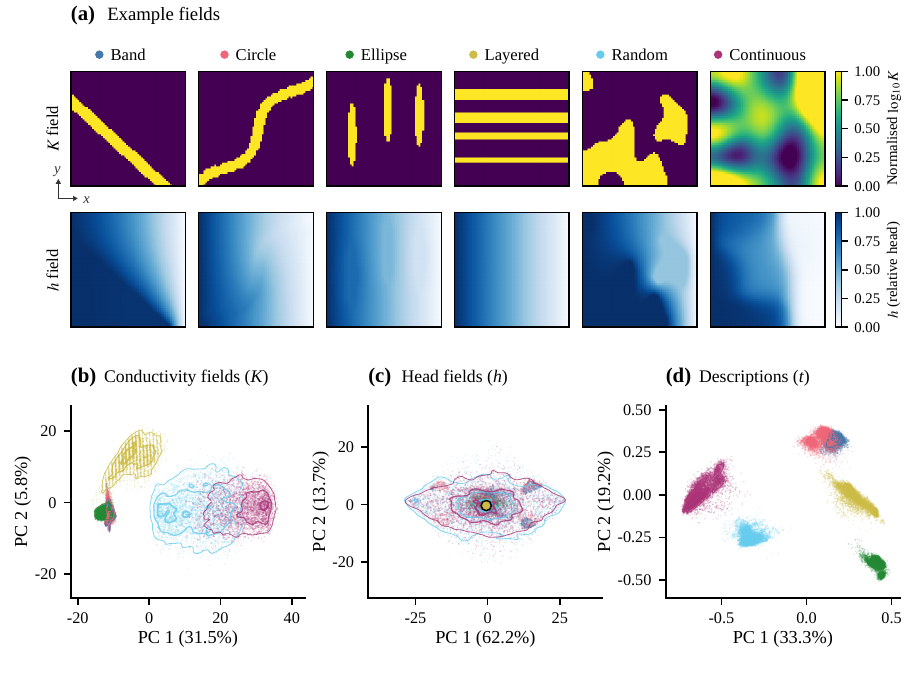}
  \caption{%
    Example fields and class structure of the synthetic dataset.
    (a)~One example realisation per class, with normalised $\log_{10}K$ above and relative hydraulic head $h$ below.
    All head fields in (a) and (c) use the horizontal-flow boundary conditions of \S\ref{sec:setting_problem}.
    Principal-component projections fitted separately to (b)~the normalised $\log_{10}K$ fields of the $18{,}000$ realisations, (c)~their head fields $h$, and (d)~the unit-norm embeddings of the $54{,}000$ descriptions $t$.
    All three splits are shown, with colours indicating class as in (a).
    In (c), the outlined marker shows the single projection shared by all Layered head fields.}
  \label{fig:dataset_overview}
\end{figure}

\begin{table}[tb]
  \centering
  \caption{The six synthetic field types treated as classes in the dataset.}
  \label{tab:classes}
  \footnotesize
  \begin{tabular}{@{}l>{\raggedright\arraybackslash}p{0.52\textwidth}>{\raggedright\arraybackslash}p{0.31\textwidth}@{}}
    \toprule
    Class & Structure & Field-generation parameters \\
    \midrule
    Band & One inclined high-$K$ sand stripe of fixed width $0.10$\,m crossing a clay background & Dip; centre height \\
    Circle & A connected sandy conduit traced by a chain of overlapping discs of radius $0.05$\,m & Mean bearing; sinuosity; lateral offset \\
    Ellipse & Three non-merging elongated sand lenses in a clay background & Orientation (horizontal/vertical); lens centres; shared axis lengths \\
    Layered & One to four horizontal sand strata in a clay background & Stratum count; centres; thicknesses \\
    Random & Spatially correlated binary field obtained by thresholding smoothed Gaussian noise & Correlation scale; clay fraction \\
    Continuous & Field constructed in $\log_{10}K$ from smoothed Gaussian noise, clipped to $[-8,-5]$, and exponentiated with base 10 & Correlation scale; pre-clipping log-mean \\
    \bottomrule
  \end{tabular}
\end{table}

The dataset comprises six types of synthetic conductivity fields, each treated as a class: four geometry-controlled types motivated by recurring sedimentary structures (inclined bands, channel-like conduits, lenses, and layers) and two geostatistical types generated from smoothed Gaussian noise (a thresholded binary field and a continuous field in log space).
These field types are simplified representations of subsurface heterogeneity and do not constitute a standard geotechnical classification.
All classes except Continuous are binary, with clay at $K = 10^{-8}$\,m/s and sand at $K = 10^{-5}$\,m/s; the Continuous field is clipped to the same bounds.
Figure~\ref{fig:dataset_overview}(a) shows an example conductivity field and its corresponding hydraulic-head field for each class.

\begin{samepage}
Let $\theta_{\mathrm{gen}}$ denote the class and its field-generation parameters (Table~\ref{tab:classes}), sampled as described in \ref{app:latent_partition}.
\begin{equation}\label{eq:dataset_generation}
\begin{aligned}
  \text{field path:}& \quad K = F(\theta_{\mathrm{gen}},\eta_K), & h &= \mathcal{D}(K), \\
  \text{text path:} & \quad \mathbf{a} = Q(\theta_{\mathrm{gen}}), & t &= T(\mathbf{a},\eta_t).
\end{aligned}
\end{equation}
Here $F$ constructs the conductivity field, with $\eta_K$ the additional random-field draw used only for Random and Continuous.
The operator $\mathcal{D}$ solves Equation~\ref{eq:darcy} under the horizontal- or vertical-flow boundary conditions of \S\ref{sec:setting_problem}.
The map $Q$ selects, summarises, and rounds the prompt attributes; $T$ generates a description, with $\eta_t$ representing wording and style variation (\S\ref{sec:setting_text}).
For Band, Circle, Ellipse, and Layered, $\theta_{\mathrm{gen}}$ determines $K$ under the fixed rasterisation (deterministic field generation), whereas for Random and Continuous it specifies distributional parameters.
\end{samepage}

We generate $18{,}000$ distinct conductivity fields, with $3{,}000$ per class.
Conductivity and head fields are represented on a $64 \times 64$ grid of nodes spanning $\Omega$, including the boundary nodes.
For horizontal-flow Layered fields, we evaluate the analytical solution of \S\ref{sec:setting_problem} on this grid.
For all other cases, we solve Equation~\ref{eq:darcy} in single precision using a five-point finite-difference discretisation with harmonic-mean interface conductivities.
Zero-normal-flux boundaries are implemented with first-order one-sided differences.
We split the realisations within each class in an $80/10/10$ ratio, yielding training, validation, and test sets of $14{,}400$, $1{,}800$, and $1{,}800$ realisations, respectively.
Each conductivity field and its three descriptions (\S\ref{sec:setting_text}) are kept together in the same training, validation, or test set.

\subsection{Geological Descriptions and Class Separation}
\label{sec:setting_text}

For each realisation, class-specific prompts supply the selected attributes $\mathbf{a}=Q(\theta_{\mathrm{gen}})$ (Equation~\ref{eq:dataset_generation}) as the only realisation-specific geological input.
From these prompts, OpenAI's \texttt{gpt-4o-mini} model \cite{openai2024gpt4omini} independently generates one canonical description and two paraphrases.
Each description is a single sentence of 20--40 words that explicitly states the supplied numerical attributes at the prescribed precision and passes rule-based checks of material and geometric consistency.
For example, an Ellipse description gives the group centroid of the three lenses but omits their individual centres, whereas a Layered description gives the centre height and thickness of every stratum (\ref{app:text_prompts}).
The prompt attributes $\mathbf{a}$ need not uniquely determine $K$, even for the four classes whose $\theta_{\mathrm{gen}}$ does.
The resulting $54{,}000$ descriptions are encoded as unit-norm, 384-dimensional vectors $\mathbf{e}$ by the frozen Sentence-BERT (SBERT) encoder \texttt{sentence-transformers/all-MiniLM-L6-v2} \cite{reimers2019sentencebert}.

Figure~\ref{fig:dataset_overview}(b--d) shows separate two-dimensional principal-component projections of the normalised log-conductivity fields, head fields, and description embeddings.
In conductivity space, Band, Circle, and Ellipse overlap strongly; Layered is visually distinct; and Random and Continuous form broader, partially overlapping distributions.
Class separation is weaker in the head fields and stronger in the description embeddings.
The Layered head fields coincide at a single point in the projection because they share the same analytical solution.
We compute silhouette scores using Euclidean distance in the original representation spaces on a seeded random subset of $3{,}000$ samples drawn separately for each representation.
The scores are $0.050$, $-0.167$, and $0.548$ for conductivity fields, head fields, and description embeddings, respectively.

\section{Methods}
\label{sec:methods}

\subsection{Solver Architecture and Input Representation}
\label{sec:methods_arch}

\begin{figure}[tbp]
  \centering
  \includegraphics[width=\linewidth]{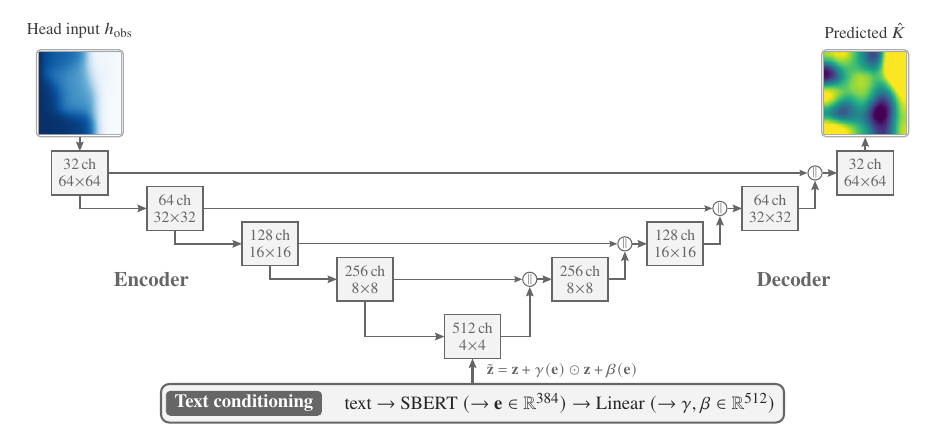}
  \caption{%
    Solver architecture.
    A U-Net maps the head input $h_{\mathrm{obs}}$ to the conductivity estimate $\hat{K}$.
    Horizontal skip connections carry encoder features to the decoder, where $\Vert$ denotes channel-wise concatenation with upsampled features.
    The description modulates only the $4 \times 4$ bottleneck via FiLM (Equation~\ref{eq:film}).
  }
  \label{fig:pipeline}
\end{figure}

The solver $G_\phi$ of \S\ref{sec:setting_problem} is a U-Net encoder--decoder \cite{ronneberger2015unet}: the observation input $h_{\mathrm{obs}}$ enters the convolutional path (one channel, $64 \times 64$), the description $t$ enters at the bottleneck, and the output is the conductivity estimate $\hat{K}$ on the same grid (Figure~\ref{fig:pipeline}).
Each convolutional block consists of two $3 \times 3$ convolutions, each followed by batch normalisation and ReLU.
The four encoder blocks (channel progression $1 \to 32 \to 64 \to 128 \to 256$), each followed by a $2 \times 2$ max-pooling, lead to a 512-channel $4 \times 4$ bottleneck block; the decoder mirrors the encoder, with each stage comprising a $2 \times 2$ stride-2 transposed convolution, a concatenation with the corresponding skip connection, and a convolutional block.
A final $1 \times 1$ convolution with a sigmoid produces the normalised log-conductivity estimate.

The description $t$ provides instance-specific side information; the solver does not represent an explicit prior distribution.
For the sentence-conditioned model (Sentence), the description enters only through Feature-wise Linear Modulation (FiLM; \cite{perez2018film}) of the bottleneck:
\begin{equation}\label{eq:film}
  \tilde{\mathbf{z}} = \mathbf{z} + \gamma(\mathbf{e}) \odot \mathbf{z} + \beta(\mathbf{e}),
\end{equation}
where $\mathbf{z} \in \mathbb{R}^{512 \times 4 \times 4}$ is the bottleneck feature map, $\mathbf{e} \in \mathbb{R}^{384}$ is the unit-norm sentence embedding of \S\ref{sec:setting_text}, and the channel-wise $(\gamma, \beta)$, each in $\mathbb{R}^{512}$ and broadcast over the $4 \times 4$ grid, are produced by a single linear layer ($384 \to 2 \times 512$).
The layer's weights and bias are initialised to zero, so every run starts at exactly the identity modulation.

In the sparse-observation regime, the input is the head sampled on an endpoint-inclusive $k \times k$ grid, $k \in \{2, 4, 8, 16, 32, 64\}$, and bilinearly interpolated back to the full grid.
Observation density therefore enters as degradation of the same one-channel input, with no observation mask and no $k$ supplied, and $k = 64$ recovers the complete field.
Results quote each grid by its point count $N=k^2$, including points on the prescribed-head boundaries.
The sparsest grid is degenerate by construction: $k = 2$ samples only the four domain corners, whose heads the boundary conditions prescribe, so its interpolated input is the same linear ramp for every realisation under a given flow direction and carries no realisation-specific information.
Throughout, the conductivity is learned and scored in a normalised log space: $\log_{10} K$ is min--max scaled to $[0,1]$ with bounds fixed a priori at the clay and sand values ($10^{-8}$ and $10^{-5}$\,m/s).
Reconstruction error is measured by the pixelwise mean squared error (MSE) between $\hat{K}$ and $K$ in this space.

\subsection{Matched Comparisons and Training Protocol}
\label{sec:methods_arms}

We first ask whether a description adds reconstruction value beyond the observed head.
Sentence uses the 384-dimensional sentence embedding of each realisation's own description, whereas the matched no-text retrain uses $\mathbf{e}=\mathbf{0}$ throughout training and evaluation while retaining FiLM.
Because descriptions also identify the field class, comparing Sentence with the no-text retrain alone cannot establish the value of instance-specific content.
We therefore add a class-identity control (the one-hot control) that supplies the true class as a six-dimensional one-hot code.
A numerical-input model, referred to as Numeric facts, augments this code with 26 attribute slots and their binary presence indicators, giving a 58-dimensional input.
Comparing the one-hot control with the no-text retrain assesses the value of class identity; comparing Sentence and Numeric facts with the one-hot control assesses the additional value of stated attributes under each representation.
Comparing Sentence with Numeric facts assesses reconstruction accuracy when the same stated facts are supplied through sentence encoding or as direct numerical input.
We construct the numerical input from the metadata used to generate each description, verify the attribute values against the stored text, and match their printed precision.
The full encoding is specified in \ref{app:numeric_facts_encoding}.

All four models share the same U-Net architecture and reconstruction-only objective.
The FiLM projection takes 384-dimensional inputs for Sentence and the no-text retrain, 6 for the one-hot control, and 58 for Numeric facts.
Each model was trained from scratch ten times, once per seed.
At each seed, all models were initialised with identical U-Net encoder--decoder weights.
The training budget was 150 epochs for Sentence and 450 for the other three models.
All models used Adam ($\beta_1=0.5$, $\beta_2=0.999$), batch size 64, initial learning rate $2\times10^{-4}$, and ReduceLROnPlateau (factor $0.5$, patience 15, floor $10^{-7}$).
In the sparse regime, training inputs were augmented by applying grid degradation to each batch with probability $0.7$, with each sample drawing $k$ uniformly from $\{2,4,8,16,32,64\}$; evaluation scored each $k$ separately.
A dense-trained Sentence control, used in the text-edit population summaries (\S\ref{sec:methods_fact_level}), received the complete head field during training and evaluation; all reconstruction-accuracy comparisons used sparse-trained models.
The selected checkpoint minimised MSE on fixed, regime-matched validation inputs shared across models at a seed.
We compared reconstruction accuracy on the test set, averaging MSE over each realisation's three descriptions within each model, seed, and observation density before computing aggregate summaries.
Sentence, the no-text retrain, the one-hot control, and Numeric facts were retrained using head fields for vertical flow, while conductivity fields, descriptions, and grouped splits were unchanged for all six classes.
Unless otherwise stated, reconstruction results are reported for horizontal flow.

\subsection{Fact-Level Diagnostics}
\label{sec:methods_fact_level}

\begin{figure}[!h]
  \centering
  \includegraphics[width=\textwidth]{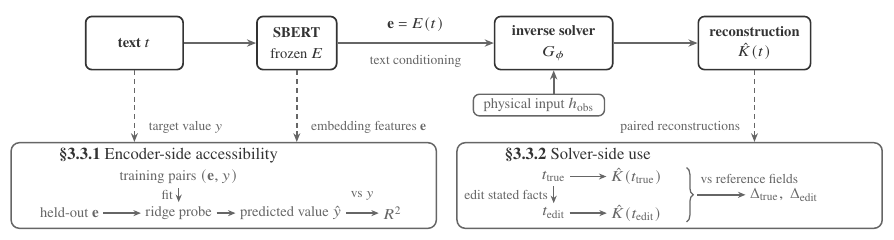}
  \caption{%
    Within-class ridge probes measure encoder-side accessibility; paired text edits test solver-side use with the physical input and solver checkpoint held fixed.
    The reported scores are probe $R^2$ and the reconstruction-error differences $\Delta_{\mathrm{true}}$ and $\Delta_{\mathrm{edit}}$ (\S\ref{sec:methods_semantic_counterfactuals}).}
  \label{fig:fact_level_diagnostics}
\end{figure}

The model comparisons of \S\ref{sec:methods_arms} treat each description as a whole and therefore cannot isolate the role of individual stated facts.
We examine these facts at two stages (Figure~\ref{fig:fact_level_diagnostics}): ridge probes measure how well these facts can be decoded from the frozen sentence embedding, and controlled text edits test whether the trained inverse solver responds to changes in those facts.

\subsubsection{Encoder-Side Accessibility of Stated Facts}
\label{sec:methods_readability}

We assess which numerical facts can be recovered from the frozen sentence embedding independently of the inverse solver.
This use of a supervised readout follows the probing framework for examining information accessible in learned representations \cite{belinkov2022probing}.
The probe targets are selected numerical attributes verbalised in the descriptions and quantities derived from them.
Linear probes match the functional form of the projection used to produce the solver's FiLM coefficients (Equation~\ref{eq:film}).
For each class and selected target attribute $y$, we fit a ridge probe $\hat y=\mathbf w^\top\mathbf e+b$ from the sentence embedding $\mathbf e$ by minimising
\begin{linenomath}
\begin{equation*}
  \sum_i (y_i-\mathbf w^\top\mathbf e_i-b)^2
  + \alpha\lVert\mathbf w\rVert_2^2,
  \qquad \alpha=1,
\end{equation*}
\end{linenomath}
over $\mathbf w$ and $b$, where $i$ indexes all training descriptions within the class and $y_i$ is, for description $i$, the stored pre-rounding value of the described attribute or a quantity derived from the stored pre-rounding attributes.
We evaluate each probe on held-out realisations from the same class to assess whether it predicts the target attribute values more accurately than the held-out class-mean baseline.
Angles are fitted as sine--cosine pairs and scored by cosine-loss $R^2_{\mathrm{circ}}$ against the held-out circular-mean baseline; other targets use ordinary $R^2$.
The visualisation in \S\ref{sec:results_readability} uses canonical validation descriptions, and the test-scored probe audit uses canonical test descriptions and both paraphrases.

\subsubsection{Solver-Side Use of Stated Facts}
\label{sec:methods_semantic_counterfactuals}

\begin{table}[htb]
  \centering
  \caption{Text edits reported in the main analysis, based on the field-generation parameters in Table~\ref{tab:classes}.}
  \label{tab:semantic_interventions}
  \footnotesize
  \begin{tabular}{@{}l
      >{\raggedright\arraybackslash}p{0.29\textwidth}
      >{\raggedright\arraybackslash}p{0.43\textwidth}@{}}
    \toprule
    Class & Edited attributes & Edit \\
    \midrule
    Band & dip angle & $+30^{\circ}$ \\
    Circle & -- & -- \\
    Ellipse & orientation and centroid & swap horizontal/vertical; transpose centroid $(x,y)\mapsto(y,x)$ \\
    Layered & all stratum centres & $+0.05$\,m \\
    Random & clay fraction & $+0.15$ \\
    Continuous & correlation length $\sigma$ & $-4$ grid cells \\
    \bottomrule
  \end{tabular}
\end{table}

In natural language processing, behavioural tests use input perturbations to examine specified changes in model predictions \cite{ribeiro2020checklist}.
Here, we test whether edits to stated facts steer the reconstruction while the frozen encoder, the solver checkpoint, and the head input $h_{\mathrm{obs}}$ are held fixed.
Each edit in Table~\ref{tab:semantic_interventions} transforms a realisation's canonical description $t_{\mathrm{true}}$ into $t_{\mathrm{edit}}$, preserving all other wording and formatting.
For example, the following Band descriptions differ only in the stated angle:

\begin{nolinenumbers}
\begin{promptbox}[title={Canonical description ($t_{\mathrm{true}}$)}]
\footnotesize\bera\setlength{\parindent}{0pt}\setlength{\baselineskip}{0.75\baselineskip}
An inclined high-permeability sand band, angled at \textbf{-5} degrees, intersects the clay background at center y = 0.60 and has a width of approximately 10 cm, creating a pronounced contrast in permeability within the geologic matrix.
\end{promptbox}
\begin{promptbox}[title={Edited description ($t_{\mathrm{edit}}$)}]
\footnotesize\bera\setlength{\parindent}{0pt}\setlength{\baselineskip}{0.75\baselineskip}
An inclined high-permeability sand band, angled at \textbf{25} degrees, intersects the clay background at center y = 0.60 and has a width of approximately 10 cm, creating a pronounced contrast in permeability within the geologic matrix.
\end{promptbox}
\end{nolinenumbers}
\noindent In some edits, retained qualitative descriptors, such as ``long-range'', conflict with the edited attributes, so the reconstruction response may reflect both the attribute change and the resulting inconsistency.
For Circle, the stated entry height and visible segment count depend on the mean bearing, so editing the bearing alone can leave the stated numerical facts mutually inconsistent.
A coordinated edit updating the bearing and these dependent quantities while preserving all other wording was not prespecified for this analysis, so Circle was not tested.

For Band, Ellipse, and Layered, we construct a reference field $K_{\mathrm{edit}}$ by applying the prescribed changes to the stored geometric parameters and regenerating the conductivity field using the original data-generation rules.
For Ellipse, we flip the orientation and transpose the centre coordinates of all three lenses.
These reference fields are used only for scoring and need not be Darcy-consistent with the fixed $h_{\mathrm{obs}}$.
Let $\hat{K}(t)$ denote the reconstruction conditioned on description $t$.
We measure the magnitude of each edit response by the root mean square (RMS) difference between $\hat{K}(t_{\mathrm{edit}})$ and $\hat{K}(t_{\mathrm{true}})$ over all pixels in normalised log-conductivity space.
For each pair $(t_{\mathrm{true}}, t_{\mathrm{edit}})$, we quantify the effect of replacing the true description with the edited description using two differences in the normalised log-$K$ MSE defined in \S\ref{sec:methods_arch}:
\begin{linenomath}
\begin{align}
  \Delta_{\mathrm{true}}
  &= \operatorname{MSE}\bigl(\hat{K}(t_{\mathrm{edit}}), K_{\mathrm{true}}\bigr)
   - \operatorname{MSE}\bigl(\hat{K}(t_{\mathrm{true}}), K_{\mathrm{true}}\bigr),
  \label{eq:delta_true}\\
  \Delta_{\mathrm{edit}}
  &= \operatorname{MSE}\bigl(\hat{K}(t_{\mathrm{true}}), K_{\mathrm{edit}}\bigr)
   - \operatorname{MSE}\bigl(\hat{K}(t_{\mathrm{edit}}), K_{\mathrm{edit}}\bigr).
  \label{eq:delta_edit}
\end{align}
\end{linenomath}
Thus, $\Delta_{\mathrm{true}}>0$ indicates increased error relative to the true field, while $\Delta_{\mathrm{edit}}>0$ indicates decreased error relative to the edited field.
For an individual Band, Ellipse, or Layered pair, positive values of both scores indicate a directional response: the reconstruction moves away from the unchanged truth and toward the edited reference.
For Random and Continuous, we do not specify an additional random-field draw $\eta_K$ for the edited parameters, so no $K_{\mathrm{edit}}$ is constructed and $\Delta_{\mathrm{true}}$ is the only field-error score.

We evaluate each sparse-trained Sentence checkpoint across the six observation grids using a fixed set of test realisations with feasible, verifiable edits.
This evaluation covers the edits in Table~\ref{tab:semantic_interventions} under horizontal flow and the Layered centre shift under vertical flow, using the Sentence checkpoints trained for each flow direction.
Population summaries (\S\ref{sec:results_interventions}) evaluate the same edits on the same realisations with the ten sparse-trained and ten dense-trained Sentence checkpoints under horizontal flow; sparse inputs follow the grid-degradation scheme of \S\ref{sec:methods_arms}, and dense inputs use the complete head field.
We also decompose the Ellipse edit by evaluating orientation-only and centroid-only edits on test data: the former changes only the orientation word (horizontally/vertically), and the latter exchanges only the stated centroid coordinates.
Both use the sparse-trained checkpoints and sparse-regime observation inputs and are scored against the combined-edit reference field $K_{\mathrm{edit}}$.
We report the ratio of the within-seed median $\Delta_{\mathrm{edit}}$ for each component edit to that for the combined edit, using the same realisations for all three edits.

\section{Results}
\label{sec:results}

\subsection{Description Conditioning across Observation Densities and Field Classes}
\label{sec:results_qualitative}

\begin{figure}[tb]
  \centering
  \includegraphics[width=\textwidth]{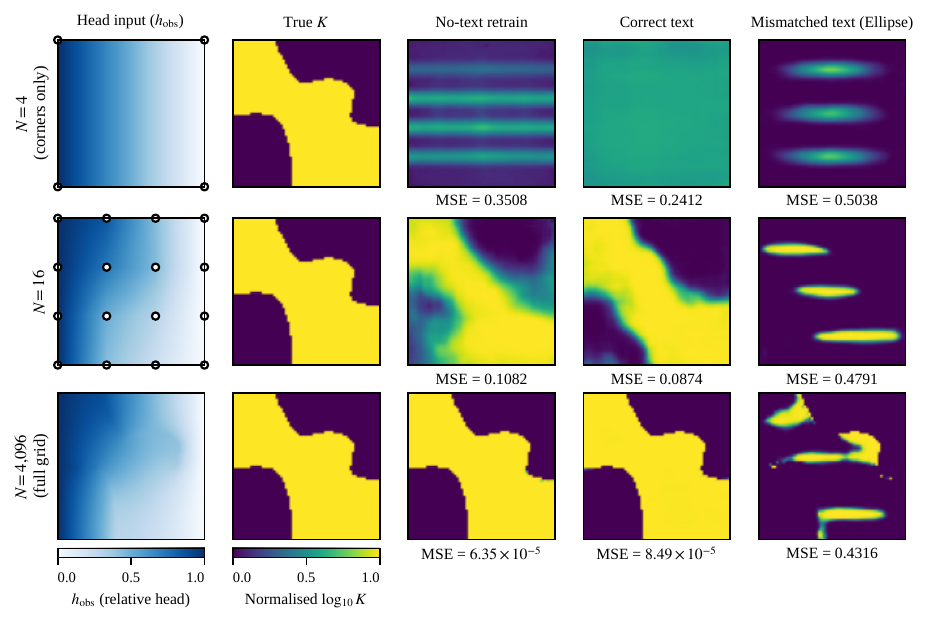}
  \caption{%
    Effect of correct and mismatched descriptions across observation densities.
    One Random validation realisation is reconstructed at the $N=4$, $16$, and $4{,}096$ observation grids by the matched no-text retrain and by Sentence, which receives either the realisation's correct text or mismatched Ellipse text naming three elongated lenses.
    All reconstructions within a row use the same head input.
    Reported MSE is against the true field.
    }
  \label{fig:text_specificity}
\end{figure}

Figure~\ref{fig:text_specificity} shows that the supplied description steers the reconstruction even when the head input is held fixed.
At the corners-only $N=4$ grid, the interpolated head is identical for every realisation (\S\ref{sec:methods_arch}) and therefore contains no realisation-specific information.
The correct Random description, which states a clay fraction and a correlation length but no patch dimensions or locations, yields a near-uniform field, whereas the mismatched Ellipse description produces the three horizontally elongated lenses it names.
The correct description also yields lower error than the matched no-text retrain at $N=4$.
At $N=16$, the head begins to constrain the coarse spatial arrangement.
The no-text retrain blurs the lower-left clay region, whereas the correct-description reconstruction recovers this region more clearly and has lower error.
Because the description states no patch locations, the improvement suggests that conditioning helps the solver use the spatial information provided by the head.
With the full head, both the no-text and correct-description reconstructions closely match the true field, although the MSE is higher with the correct description in this example.
The reconstruction with the mismatched description remains farther from the truth than the other two at all three densities shown.
Because the mismatch also changes the class, this example shows class-level semantic steering but not the use of instance-specific content; \S\ref{sec:results_arm_ladder} takes up that distinction.

\begin{figure}[tb]
  \centering
  \includegraphics[width=\textwidth]{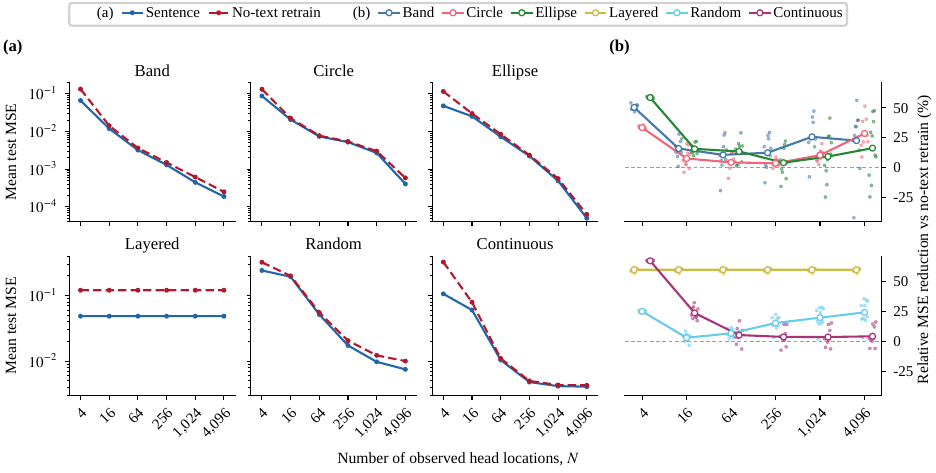}
  \caption{%
  Class-specific performance across observation density after sparse training, over ten retraining seeds.
  (a)~Mean test MSE of Sentence and the matched no-text retrain.
  Logarithmic MSE limits differ between rows.
  (b)~Paired relative MSE reduction of Sentence over the no-text retrain: small points show individual seeds, and open circles connected by lines show means.
  Both rows share the same scale; positive values favour sentence conditioning.
  In (b), points are offset horizontally for visibility.
  }
  \label{fig:perclass_ladder}
\end{figure}

The principal-component projections in Figure~\ref{fig:dataset_overview}(b,c) show weaker class separation for head $h$ than for conductivity $K$.
We therefore assess whether description conditioning improves reconstruction accuracy across field classes and observation densities (Figure~\ref{fig:perclass_ladder}).
Mean paired MSE reductions are positive for every class and observation density, but the consistency of improvement across training seeds varies.
At the full grid, both models achieve MSE below $10^{-3}$ for Band, Circle, and Ellipse, so the absolute gains from conditioning are small.
Random and Continuous retain larger errors, with mean paired reductions of $24.2\%$ and $4.2\%$, respectively.
For Layered, both models' errors remain nearly flat as observation density increases, consistent with the exact head being invariant to stratum position (\S\ref{sec:setting_problem}).
Section~\ref{sec:results_arm_ladder} examines whether this benefit comes from class identity alone or also from instance-specific content.

\FloatBarrier
\subsection{Reconstruction Value of Instance-Specific Content}
\label{sec:results_arm_ladder}

\begin{figure}[!htb]
  \centering
  \includegraphics[width=252pt]{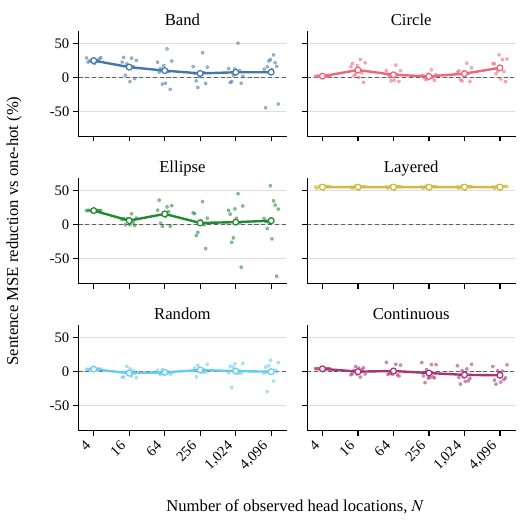}
  \caption{%
    Reconstruction value beyond class identity after sparse training.
    Paired relative MSE reduction of Sentence over the one-hot control for each class; positive values favour Sentence.
    Small points show individual seeds, and open circles connected by lines show means across ten retraining seeds.
    Small points are offset horizontally for visibility.
    }
  \label{fig:class_identity}
\end{figure}

Because descriptions identify the field class, the advantage of Sentence over the no-text retrain does not by itself establish the value of instance-specific content.
We therefore compare Sentence with the one-hot control of \S\ref{sec:methods_arms} in Figure~\ref{fig:class_identity}.
With corners-only observations, which contain no realisation-specific head information, Sentence has lower MSE than the one-hot control for all six classes in all ten seeds.
With denser observations, the mean gain over the one-hot control remains positive for Band, Circle, Ellipse, and Layered but becomes negative for Random and Continuous at some densities.
For Layered, which shows the largest mean gain at every density, neither the class label nor the head locates the strata, whereas the description states their positions.

\begin{figure}[tb]
  \centering
  \includegraphics[width=\textwidth]{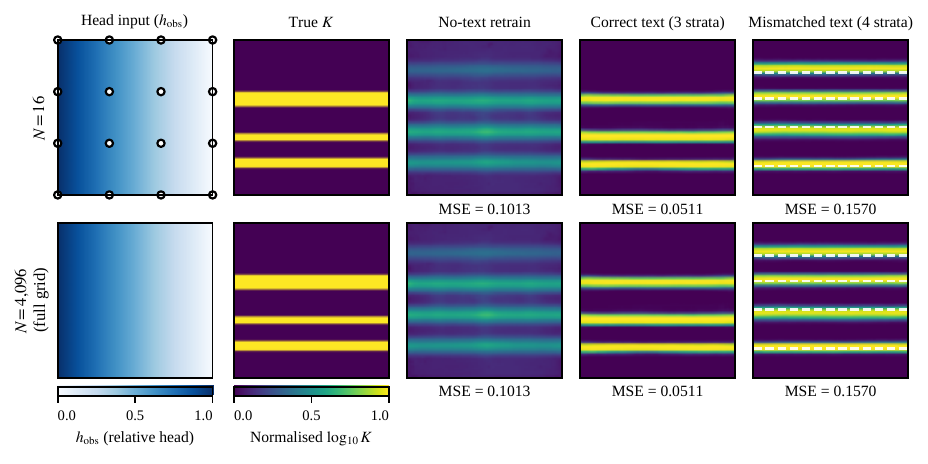}
  \caption{%
    A same-class text mismatch steers reconstruction toward the stated stratum configuration.
    One three-stratum Layered validation realisation is reconstructed at $N=16$ and $N=4{,}096$ by the matched no-text retrain and by Sentence under either the realisation's correct text or mismatched text from a four-stratum realisation of the same class.
    All reconstructions within a row use the same head input; dashed lines mark the four centres stated by the mismatched text.
    }
  \label{fig:same_class_description_mismatch}
\end{figure}

Figure~\ref{fig:same_class_description_mismatch} shows the effect of swapping descriptions within Layered under horizontal flow, complementing the cross-class mismatch in Figure~\ref{fig:text_specificity}.
With the trained Sentence model and head input fixed, the correct description of the three-stratum target is replaced by one describing another Layered realisation with four strata at different centres.
At both $N=16$ and the full grid, the correct-description reconstruction recovers the three target stratum positions, while the mismatched-description reconstruction follows the four centres stated in that text.
Accordingly, the correct description yields lower error than the no-text retrain, whereas the mismatched description yields higher error.
All three reconstructions remain essentially unchanged across these densities, consistent with denser samples of the exact head supplying no additional information about stratum position.
Because class identity is unchanged, the response to the swapped description demonstrates the use of instance-specific content.

\begin{figure}[!t]
  \centering
  \includegraphics[width=\textwidth]{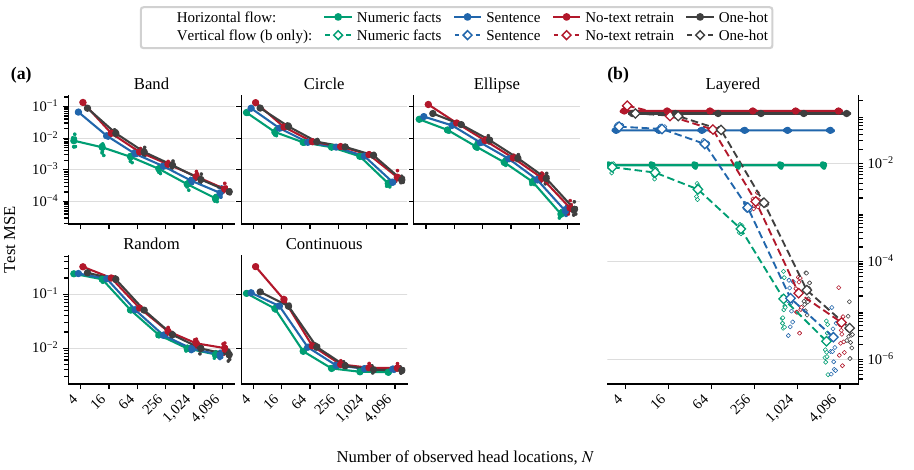}
  \caption{%
    Test MSE of the four models after sparse training.
    (a)~The five non-Layered classes under horizontal flow.
    (b)~Layered under horizontal and vertical flow.
    Small points show individual seeds; lines connect means across ten retraining seeds.
    MSE axes are logarithmic, with shared limits within each row of (a) and separate limits for (b).
    Points are offset horizontally for visibility.}
  \label{fig:numeric_facts_test_mse}
\end{figure}

We next compare Sentence with Numeric facts to examine how reconstruction changes when the same instance-specific content is supplied directly as numerical input.
Figure~\ref{fig:numeric_facts_test_mse}(a) compares test MSEs for the non-Layered classes across observation densities under horizontal flow.
Numeric facts yields lower mean MSE than Sentence for Band, Ellipse, and Continuous at every evaluated density.
For Circle, the advantage of Numeric facts is most pronounced with sparse observations.
For Random, Sentence has slightly lower mean MSE than Numeric facts at $N \geq 256$.
Figure~\ref{fig:numeric_facts_test_mse}(b) compares horizontal and vertical flow through the same Layered fields.
Under horizontal flow, reconstruction errors remain nearly constant across observation densities.
Numeric facts achieves lower MSE than Sentence in all ten seeds at each density, with a mean per-seed MSE reduction of $80.6\%$.
Under vertical flow, the head becomes sensitive to stratum position, and errors decrease sharply as observations become denser.
For Layered, Numeric facts achieves the lowest mean test MSE among all four models at every evaluated density under both flow directions.
Under vertical flow, however, the ordering of Numeric facts and Sentence varies across seeds at $N=1{,}024$ and $4{,}096$.

\FloatBarrier
\subsection{Fact Accessibility and Reconstruction Responses}
\label{sec:results_fact_pathway}

The comparisons in \S\ref{sec:results_qualitative} and \S\ref{sec:results_arm_ladder} evaluate each description as a whole.
We therefore examine whether individual stated facts are linearly accessible in the frozen embedding and whether controlled edits to those facts directionally steer the reconstruction with the head input fixed.

\subsubsection{Linear Accessibility of Stated Facts}
\label{sec:results_readability}

\begin{figure}[!t]
  \centering
  \includegraphics[width=\textwidth]{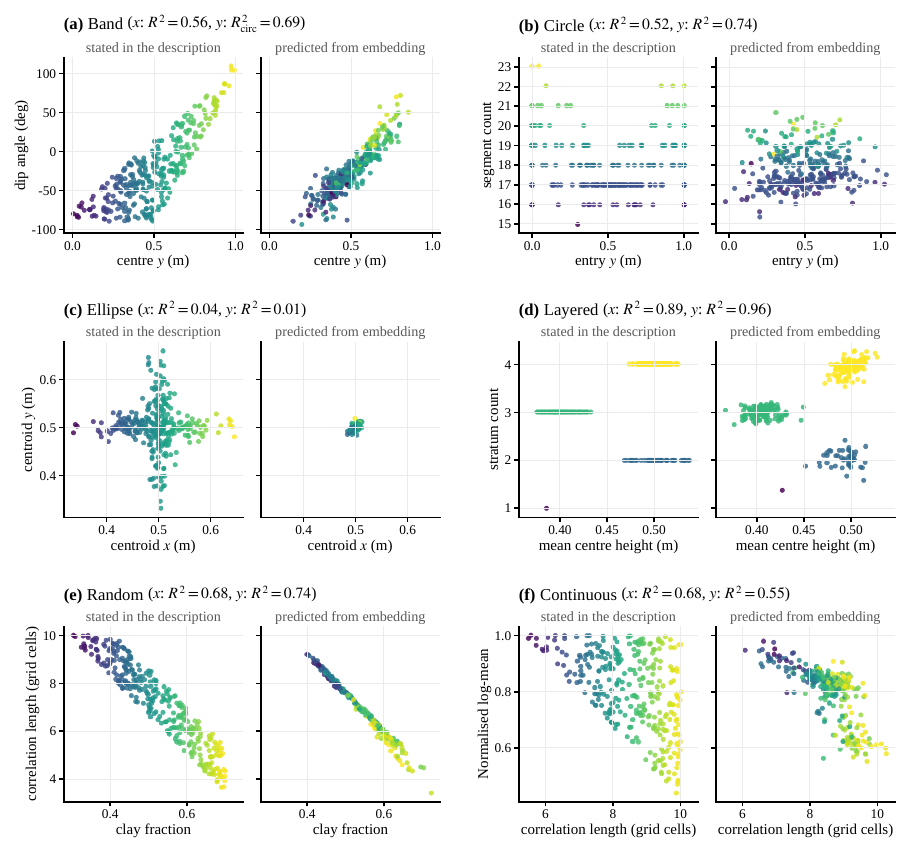}
  \caption{%
    Linear-probe decoding of numerical attributes from frozen sentence embeddings.
    For each class, the left panel shows two attributes stated in or derived from $300$ canonical validation descriptions; the right panel shows predictions from ridge probes fitted on the disjoint training split.
    Paired panels share axis limits and retain each realisation's colour.
    Colour encodes the true $x$-axis value, except in panels~(b) and~(d), where it encodes the true $y$-axis value.
    Scores are validation $R^2$, except for the Band dip angle, which uses the cosine-loss circular analogue $R^2_{\mathrm{circ}}$.
    For Continuous, the normalised log-mean is the pre-clipping parameter.}
  \label{fig:stated_vs_decoded}
\end{figure}

Figure~\ref{fig:stated_vs_decoded} visualises linear-probe decoding of two numerical attributes per class from the frozen embeddings of held-out validation descriptions.
For Band and Circle, the colour ordering remains visible after decoding, despite changes in the shape and spread of the point clouds.
The Ellipse centroid predictions cluster tightly around the class mean.
The Layered predictions preserve distinct groups corresponding to stratum count, together with the associated differences in mean centre height.
This retained count information is consistent with the reconstruction adopting the stratum count stated in the mismatched description (Figure~\ref{fig:same_class_description_mismatch}).
For Random, decoding compresses variation in the stated clay fraction and correlation length into a narrow trend, but the correlation between the two stated values leaves open whether each fact is read out directly or predicted through other correlated content.
For Continuous, the colour ordering also remains visible in the decoded point cloud.
The test-scored probe audit (\S\ref{sec:methods_readability}) shows the same differences in accessibility across attributes.
In that audit, the Ellipse major and minor axis lengths remain partly predictable despite the weak centroid readout.

\subsubsection{Reconstruction Responses to Fact Edits}
\label{sec:results_interventions}

\begin{figure}[tbp]
  \centering
  \includegraphics[width=\textwidth]{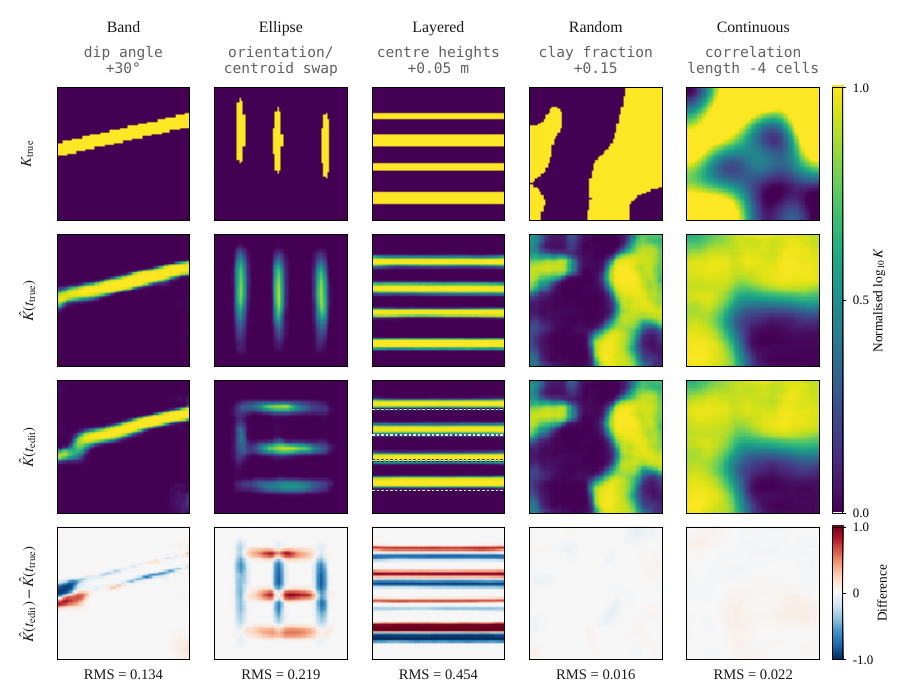}
  \caption{%
    Reconstruction responses to text edits at fixed observations ($N=16$) for the sparse-trained checkpoint.
    Rows show the true field, reconstructions conditioned on the canonical and edited descriptions, and their signed difference.
    Colour scales are shared across classes; RMS labels quantify the magnitude of the reconstruction change.
    Dashed lines in the Layered edited reconstruction mark the unedited stratum centres.}
  \label{fig:intervention_reconstructions}
\end{figure}

Linear accessibility in the frozen embedding does not establish solver-side use.
We therefore apply the text edits defined in \S\ref{sec:methods_semantic_counterfactuals} to test whether a stated-fact edit directionally steers the reconstruction while the head input is held fixed.
Figure~\ref{fig:intervention_reconstructions} shows five illustrative validation examples at $N=16$.
For Band, the increase in dip angle produces changes concentrated near one end of the stripe.
For Ellipse, the reconstructed lenses reorient from vertical to horizontal, with accompanying positional changes.
For Layered, the centre shift displaces the reconstructed horizontal interfaces upward.
Increasing the Random clay fraction and decreasing the Continuous correlation length produce only weak, diffuse changes.

\begin{figure}[tbp]
  \centering
  \includegraphics[width=252pt]{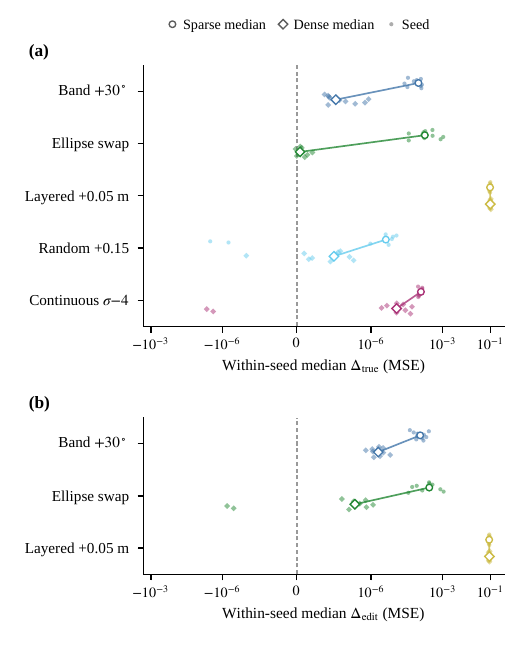}
  \caption{%
    Within-seed test medians of $\Delta_{\mathrm{true}}$ (a) and $\Delta_{\mathrm{edit}}$ (b) for the sparse-trained and dense-trained Sentence models under horizontal flow, on a shared symmetric-log scale.
    $\Delta_{\mathrm{edit}}$ is defined here only for Band, Ellipse, and Layered.
    Sparse inputs follow the grid-degradation scheme of \S\ref{sec:methods_arms}, while dense inputs use the full $N=4{,}096$ grid.
    Small filled markers show individual seeds; large open markers show the median across ten seeds.
    Circles and diamonds denote the sparse and dense regimes, respectively; line segments connect their medians.
    Markers are offset vertically within each category to reduce overlap.}
  \label{fig:intervention_population}
\end{figure}

Figure~\ref{fig:intervention_population} summarises the within-seed medians of $\Delta_{\mathrm{true}}$ and $\Delta_{\mathrm{edit}}$ over the test population for the sparse- and dense-trained Sentence models.
Under sparse training, the Band, Ellipse, and Layered edits give positive medians of both scores in every seed, supporting a directional response away from the true field and toward the edited reference.
For Random, the within-seed medians of $\Delta_{\mathrm{true}}$ are small and have mixed signs in both regimes.
Under sparse training, the across-seed median $\Delta_{\mathrm{true}}$ for Continuous is comparable to that for Band; movement toward an edited reference field is not evaluated because $K_{\mathrm{edit}}$ is not specified for Continuous.
Because many edited Continuous descriptions retain a correlation-range descriptor that conflicts with the edited value, this response cannot be attributed to the edited value alone.
With dense training and full observations, the across-seed medians of both scores for Band and Ellipse shrink by more than an order of magnitude, whereas those for Layered remain similar.
This persistence is consistent with the head field's invariance to stratum position under horizontal flow.
We also decompose the Ellipse edit under sparse training: the within-seed median $\Delta_{\mathrm{edit}}$ for the orientation edit alone ranges from $0.94$ to $1.08$ times that of the combined edit across the ten seeds.
The corresponding ratio for the centroid edit alone is $-0.0001$ to $0.0022$, indicating little change in error against the edited reference.
The centroid coordinates are also weakly decoded by the linear probes in Figure~\ref{fig:stated_vs_decoded}(c).

\begin{figure}[tbp]
  \centering
  \includegraphics[width=252pt]{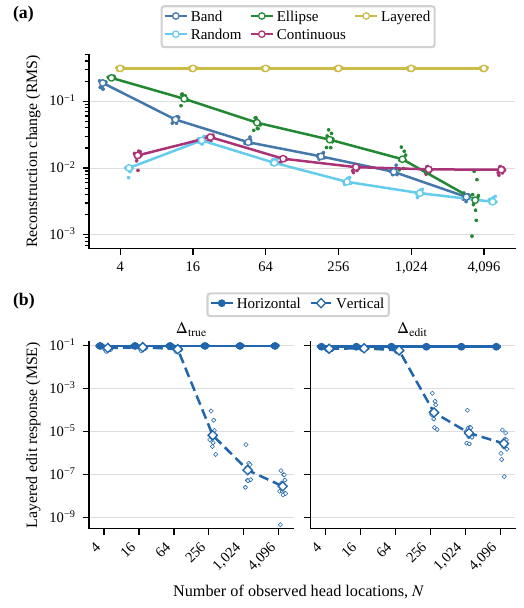}
  \caption{%
    Text-edit responses across observation densities within fixed sparse-trained checkpoints.
    (a) RMS reconstruction change for the edits in Table~\ref{tab:semantic_interventions} under horizontal flow.
    (b) $\Delta_{\mathrm{true}}$ (left) and $\Delta_{\mathrm{edit}}$ (right) for the Layered centre shift under horizontal and vertical flow.
    Small points show within-seed test medians; large points and lines show medians across ten seeds.
    Points are offset horizontally for visibility.}
  \label{fig:edit_density_response}
\end{figure}

Figure~\ref{fig:edit_density_response} shows how reconstruction responses to text edits vary with observation density within each sparse-trained checkpoint.
In panel~(a), the Band and Ellipse RMS reconstruction changes decrease with observation density under horizontal flow, while the Layered response remains nearly constant.
The Random and Continuous median curves first rise and then decline, with Continuous levelling off at high density.
In panel~(b), the across-seed medians of both directional scores for Layered remain positive and nearly constant under horizontal flow, but under vertical flow they fall sharply between $N=64$ and $N=256$.
These results suggest that denser observations attenuate text-edit responses when the head field is informative about the edited geometry.

\section{Discussion}
\label{sec:discussion}

The synthetic experiments with idealised descriptions allowed us to examine the value of geological descriptions while controlling both their factual content and the physical observation conditions.
Descriptions can improve reconstruction where head observations leave structural ambiguity, and selected stated facts differ in their linear accessibility and in how editing them affects reconstruction.
Given the same stated facts, Numeric facts achieves lower mean MSE than Sentence in most reported conditions.

\subsection{Information from Head Observations and the Role of Geological Descriptions}
\label{sec:disc_head_information}

The role of a geological description depends in part on how informative the head observations are about the described structure.
Flow direction determines whether the head field responds to that structure, while sampling density determines how much of the field is observed.
Under horizontal flow through Layered fields, the exact head solution is invariant to stratum position (\S\ref{sec:setting_problem}).
Bilinear interpolation reproduces the linear head at every evaluated density, so the solver input is unchanged up to numerical precision.
Reconstruction errors remain nearly flat across observation density, while Sentence retains lower error than the no-text retrain and the one-hot control (\S\ref{sec:results_arm_ladder}).
The response to the stated centre shift is consistent with the solver using descriptions to resolve ambiguity that the complete head field still leaves (Figure~\ref{fig:edit_density_response}(b)).

We would therefore expect a weaker response to the stated centre shift when head observations become more informative about stratum position.
Under vertical flow through the same Layered fields, the head is sensitive to stratum position, and both the mean reconstruction error of Sentence and its centre-edit response decrease sharply from $N=64$ to $N=256$ (Figures~\ref{fig:numeric_facts_test_mse}(b) and~\ref{fig:edit_density_response}(b)).
The Band and Ellipse edit responses also weaken as observation density increases under horizontal flow (Figure~\ref{fig:edit_density_response}(a)), although not every evaluated class shows a monotonic decrease with density.
These results support the interpretation that the solver relies less on the stated geometry of selected features when head observations become more informative about those features.
Even when responses to selected edits weaken, the complete description can retain an accuracy benefit.
In the vertical-flow Layered case, for example, Sentence retains lower mean reconstruction error than the no-text retrain and the one-hot control at every evaluated density, although the ordering varies across seeds at $N \geq 256$.

Under horizontal flow and sparse training on this synthetic dataset, Sentence achieves positive mean paired relative MSE reductions over the matched no-text retrain in all six classes and at every evaluated density, not only in the classes with deterministic field generation (Figure~\ref{fig:perclass_ladder}).
Paired gains change sign across seeds in some conditions, for reasons that remain unresolved.
The absolute gains are small where both models already reconstruct accurately.
Relative to the one-hot control, at the evaluated densities with $N \geq 16$, mean paired relative MSE reductions are positive at every density for Band, Circle, Ellipse, and Layered but negative at most densities for Random and Continuous (Figure~\ref{fig:class_identity}).
For these two classes, the description specifies distributional parameters rather than the realised field (\S\ref{sec:setting_fields}).

\subsection{Representation and Use of Stated Facts}
\label{sec:disc_fact_representation}

The known field-generation attributes allow the same stated facts to be supplied as sentences or numerical inputs.
Numeric facts gives lower mean MSE than Sentence in most conditions shown in Figure~\ref{fig:numeric_facts_test_mse}; the exception is Random at $N \geq 256$, where the mean MSEs differ little.
Under vertical flow, which Figure~\ref{fig:numeric_facts_test_mse} shows only for Layered, Sentence has lower mean MSE than Numeric facts at some denser observation grids for Band, Circle, and Ellipse.
Because the two models also differ in input dimensionality and training budget, the observed differences in MSE cannot be attributed solely to sentence encoding.
For Layered, the fact-level diagnostics show that the stratum count and mean centre height are linearly predictable (Figure~\ref{fig:stated_vs_decoded}(d)), and that the centre-shift edit gives positive within-seed medians of both directional scores under horizontal flow and sparse training (Figure~\ref{fig:intervention_population}).
Nevertheless, under horizontal flow, Numeric facts achieves lower MSE than Sentence in all ten seeds at each evaluated density (Figure~\ref{fig:numeric_facts_test_mse}(b)).
Linear accessibility and directional edit responses therefore do not establish comparable reconstruction accuracy under sentence and numerical conditioning.

Numerical decoding has been studied at the token level, with accuracy varying across embedding methods \cite{wallace2019numeracy}.
In the present study, we assess attributes within complete geological descriptions and find differences in both their linear accessibility and their influence on reconstruction.
For Ellipse, editing the orientation word alone gives a within-seed median $\Delta_{\mathrm{edit}}$ close to that of the combined edit, whereas exchanging the stated centroid coordinates alone changes the error against the common combined-edit reference very little (\S\ref{sec:results_interventions}).
The centroid is also weakly decoded by the linear probes, while the axis lengths remain partly predictable (\S\ref{sec:results_readability}).
The stated centroid is the group centroid of the three lenses and varies over a narrow range (Figure~\ref{fig:stated_vs_decoded}(c)), and individual lens centres are not stated (\ref{app:text_prompts}).
The contrast between the orientation and centroid edits therefore does not separate the form in which an attribute is expressed from how strongly it constrains the field.
Other numerically stated attributes behave differently: edits to the Band dip angle and the Layered stratum centres give positive within-seed medians of both directional scores under sparse training.
For Random, the clay fraction is partly predictable from the embedding, yet editing it produces only small median changes in reconstruction error.
Linear probes and text edits provide complementary diagnostics, reflecting the distinction between extractability and use discussed in probing studies \cite{belinkov2022probing}.
The accuracy benefit of individual facts remains to be quantified.

\subsection{Implications for Using Geological Knowledge}
\label{sec:disc_observation_design}

Descriptions of geological facts can help reconstruct structure that head observations leave unresolved, as in the horizontal-flow Layered case, where the exact head is invariant to stratum position.
This invariance means that the exact head also provides no information for verifying the stated stratum positions; consistent with this, the centre-shift edit produces a positive median increase in error against the true field even when the complete head field is observed (\S\ref{sec:disc_head_information}).
The solver's responses differ across edits of selected stated facts, and an accuracy gain from a complete description alone does not identify which facts it used.
Using geological descriptions in inversion thus calls for three separate assessments: what information a description adds to the physical observations, how the solver uses that information, and how reliable the description is.
Separating these assessments is consistent with the call to address trustworthiness explicitly in geotechnical research and practice \cite{phoon2024trustworthy}.

Beyond the Darcy setting, these findings motivate broader evaluation of language-based side information in ill-posed inverse problems, building on existing work on semantic regularisation \cite{zhang2024semantic,chen2024semanticem}.
In geotechnical site characterisation, prior geological knowledge has also been incorporated by using a geological cross-section from a previous project with similar geological settings as a training image, together with limited site-specific boreholes \cite{shi2021icxgboost}.
Motivated by the response to the Ellipse orientation word (\S\ref{sec:disc_fact_representation}), future studies could test whether qualitative structural descriptions help distinguish solutions consistent with the measurements.
Another direction is to compare sentence conditioning with conditioning on explicit attributes or constraints derived from the same descriptions; the numerical input used here, constructed from simulator metadata and verified against the descriptions (\S\ref{sec:methods_arms}; \ref{app:numeric_facts_encoding}), provides a reference case of exact recovery of the encoded attributes at their stated precision.
Language models could assist in constructing these representations from existing records or expert statements \cite{wu2024futureproofing,qian2025llm}, with their contribution evaluated against simpler extraction methods using extraction accuracy, verification effort, and reconstruction accuracy.

Evaluations should distinguish agreement with the source text from independent support for the stated knowledge.
To isolate the effects of information availability and precision, learned solvers should be trained and evaluated under matching information conditions.
For selected geometric features, edit responses weakened as the head observations became more informative; a further direction is to represent uncertainty in the supplied statements explicitly and to adapt their influence to source reliability and to the information provided by the measurements.
Environmental modelling tools such as pyEMU support data-worth analyses within a framework for parameter and prediction uncertainty analysis \cite{white2016pyemu}.
Information entropy has been used to select efficient sampling locations for geotechnical site characterisation, including additional locations at later stages of an investigation \cite{zhao2019sampling}.
Future evaluations could extend these approaches to site investigation planning that covers both measurements and domain knowledge, comparing the expected reconstruction benefit of an additional measurement with that of acquiring or verifying a geological statement, accounting for reliability and cost.

\subsection{Scope and Limitations}
\label{sec:disc_limitations}

The idealisation that enables these comparisons also bounds their scope.
The synthetic dataset uses idealised descriptions that state known simulator attributes and pass rule-based consistency checks (\S\ref{sec:setting_text}).
Transfer to site records, such as borehole logs and investigation reports, that contain incomplete, imprecise, or conflicting geological information remains untested.
The reported experiments also use synthetic head observations generated under prescribed Darcy physics and boundary conditions, without added measurement noise.
They therefore do not establish robustness to measurement noise or uncertainty in the governing physics or boundary conditions.
The conductivity fields are two-dimensional cross-sections on the unit-square domain of \S\ref{sec:setting_problem}, so the results do not address three-dimensional heterogeneity or site-scale domains.

The numerical comparison uses inputs constructed from metadata; an advantage of natural-language encoding or a benefit specific to language models has not been established.
The fact-level diagnostics are limited to selected attributes and edits, without systematically assessing interactions among facts.
Linear readout cannot establish that a poorly decoded fact is absent from the embedding \cite{belinkov2022probing}.
Whether these diagnostics predict the reconstruction benefit of acquiring individual facts or combinations of facts, and whether the findings generalise across geological structures, inverse problems, text generators, sentence encoders, and solver architectures, remain to be tested.

\section{Conclusion}
\label{sec:conclusion}

This study examined the contribution and limitations of geological descriptions as side information for learned conductivity inversion, using a synthetic Darcy-flow dataset of two-dimensional cross-sections with idealised descriptions and no added measurement noise.
After sparse-observation training, geological descriptions can improve conductivity reconstruction over a matched solver retrained without descriptions where head observations leave structural ambiguity.
A comparison with the one-hot control showed that instance-specific content can contribute beyond field-type information.
Using linear probes and controlled text edits, we found that selected stated facts differed in their linear accessibility and in their influence on reconstruction.
Editing the stated geometry of selected features shifted reconstructions less strongly as head observations became more informative about those features.
In the horizontal-flow Layered case, where the head is invariant to stratum position, shifting the stated positions produced a positive median increase in reconstruction error even with complete head observations, showing that accuracy remained sensitive to errors in those positions.
Supplying the stated facts as numerical inputs gave lower mean MSE than sentence conditioning in most reported conditions, although input dimensionality and training budget also differed; an advantage of natural language over a structured representation of the same facts was therefore not established.

Future research should evaluate language-based side information in other ill-posed inverse problems, assessing separately what descriptions add to measurements, how solvers use that information, and how reliable the descriptions are.
Such studies should compare sentence conditioning against conditioning on structured representations derived from the same sources, and assess sensitivity to incomplete, imprecise, or inaccurate descriptions and to measurement noise.
Researchers could use these evaluations to plan which physical measurements to collect and which statements to acquire or verify, accounting for expected reconstruction benefit, reliability, and cost.

\begin{samepage}
\section*{Data availability}

The synthetic dataset, fixed splits for training, validation and testing, data for controlled text edits, and accompanying data-generation and loading code are available on Zenodo at \url{https://doi.org/10.5281/zenodo.22722855} \cite{darcy_dataset}.

\end{samepage}

\section*{Declaration of generative AI and AI-assisted technologies in the manuscript preparation process}

During the preparation of this work, the authors used Claude (Anthropic) in order to draft, revise and proofread the text. After using this tool, the authors reviewed and edited the content as needed and take full responsibility for the content of the published article.

\appendix
\setcounter{table}{0}
\counterwithin*{table}{section}
\setcounter{figure}{0}

\section{Class assignment and parameter mapping}
\label{app:latent_partition}

\begin{table}[htbp]
\centering
\caption{Coordinates controlling the field-generation parameters of each field class.}
\label{tab:latent_mapping}
\footnotesize
\begin{tabular}{ll}
  \toprule
  Class & Coordinates controlling field-generation parameters \\
  \midrule
  Band       & $\xi_1$ (dip angle), $\xi_1, \xi_2$ (centre height $c_y$) \\
  Circle     & $\xi_3$ (mean bearing), $\xi_4$ (sinuosity), $\xi_5$ (lateral offset) \\
  Ellipse    & $\xi_3$ (orientation), $\xi_4$--$\xi_6$ (longitudinal centres), $\xi_7$--$\xi_9$ (transverse jitter), $\xi_{10}$--$\xi_{11}$ (axes) \\
  Layered    & $\xi_3$ (stratum count), $\xi_4$--$\xi_7$ (placement), $\xi_8$--$\xi_{11}$ (thicknesses) \\
  Random     & $\xi_1, \xi_2$ (correlation scale, clay fraction) \\
  Continuous & $\xi_1, \xi_2$ (correlation scale, log-mean) \\
  \bottomrule
\end{tabular}
\end{table}

We construct $\theta_{\mathrm{gen}}$ from candidate eleven-dimensional vectors $\boldsymbol{\xi} = (\xi_1, \dots, \xi_{11})$ with independent $\mathrm{U}[0,1]$ components.
We divide the range of $\tau = \xi_1 + \xi_2$ into six intervals with approximately equal probability under this candidate distribution.
The intervals, ordered by increasing $\tau$, correspond to Band, Circle, Ellipse, Layered, Random, and Continuous.
The generator draws $\boldsymbol{\xi}$ conditional on each class by rejection sampling and rejects exact duplicate conductivity rasters until it retains $3{,}000$ distinct fields per class.
Rejecting duplicate rasters leaves $33$, $505$, $1{,}212$, and $1{,}250$ Layered fields with one to four strata, respectively.

Table~\ref{tab:latent_mapping} lists the coordinates controlling the field-generation parameters in $\theta_{\mathrm{gen}}$.
For Band, the centre height is $c_y = \mathrm{reflect}_{[0,1]}\!\big(0.5 + (\xi_1 - \xi_2)\big)$, where $\mathrm{reflect}_{[0,1]}$ folds values into $[0,1]$ by reflection at the boundaries.
For Random and Continuous, the generator spatially smooths Gaussian noise $\eta_K$ (Equation~\ref{eq:dataset_generation}).
The correlation scale $\sigma$ is the standard deviation of the Gaussian smoothing kernel, measured in grid cells; the descriptions call it ``correlation length''.
Random thresholds the smoothed field at a rank selected by the clay fraction.
Continuous standardises the smoothed field, applies the prescribed log-mean and fixed log-spread ($1.5$ in $\log_{10}$ units), clips to the conductivity bounds in log space, and exponentiates.
The class partition constrains the joint distribution of Band dip and centre height and restricts the parameter ranges of Random and Continuous.
Within Random and Continuous, the correlation scale is negatively correlated with the clay fraction and log-mean, respectively.
The complete parameter mappings are provided in the generation code accompanying the dataset.

\section{Text generation prompts and paraphrase protocols}
\label{app:text_prompts}

For each of the $18{,}000$ realisations, \texttt{gpt-4o-mini} (temperature $1.0$, \texttt{max\_tokens} $=100$) independently generates one canonical description and two paraphrases from the same generator attributes.
These $54{,}000$ descriptions form the conditioning dataset.
Each user prompt combines a class-specific template with a style hint and fidelity instructions requiring the selected attributes to be stated explicitly.
Paraphrase requests also include the additional instruction shown below.

\begin{nolinenumbers}
\begin{promptbox}[title={System prompt (full text; shared across classes and roles)}]
\footnotesize\bera\setlength{\parindent}{0pt}\setlength{\baselineskip}{0.75\baselineskip}
You are a geotechnical engineer describing subsurface permeability fields in a 1m x 1m cross-section.
Write exactly ONE concise sentence (20-40 words) using geotechnical vocabulary.\\
IMPORTANT RULES:\\
1. Begin directly with the specific geological feature {\fontfamily{lmr}\selectfont\textemdash} never with generic phrases like \textquotesingle The subsurface\textquotesingle, \textquotesingle The permeability\textquotesingle, or \textquotesingle A spatially correlated\textquotesingle.\\
2. VARY your vocabulary and sentence structure for EVERY description, but use direct, literal scientific wording. Prefer precise verbs such as crosses, occupies, lies within, is embedded in, or forms. Avoid decorative, metaphorical, ornate, or vague language. Never use the same opening phrase twice.\\
3. Output only that sentence: no preface, explanation, list, alternative, or second sentence. Stop after 40 words.\\
4. Do not invent boundary conditions or a flow direction. Never describe groundwater or flow as upward, downward, or vertical; geometric features may still be vertically oriented.\\
5. The domain is a vertical cross-section, not a map view. Never use compass directions (north, south, east, west) for any feature. Use horizontal, vertical, inclined, left, or right instead.\\
6. Vertical positions are heights measured upward from the lower boundary. Never call a height a depth.
\end{promptbox}

\begin{promptbox}[title={Core user-prompt templates}]
\footnotesize\bera\setlength{\parindent}{0pt}\setlength{\baselineskip}{0.75\baselineskip}
\textbf{Band.}
Describe an inclined high-permeability sand band in a low-permeability clay background.\\
- Orientation: $\langle$angle-category$\rangle$ (approximately $\langle$angle$\rangle$ degrees from horizontal)\\
- Position: passes through the $\langle$region$\rangle$ of the domain (center y = $\langle y\rangle$)\\
- Band width: approximately $\langle w\rangle$ cm\\[5pt]
\textbf{Circle.}
Describe a connected sandy conduit (high-permeability pathway) in a clay matrix.\\
- Mean bearing: $\langle d\rangle$ degrees ($\langle$direction-category$\rangle$); write the figure $\langle d\rangle$ exactly, with no decimal point\\
- Path character: $\langle$curvature-category$\rangle$ (curvature parameter = $\langle c\rangle$)\\
- Entry height: $\langle$region$\rangle$ of the domain (y = $\langle y\rangle$)\\
- Composed of approximately $\langle n\rangle$ connected permeable segments\\[5pt]
\textbf{Ellipse.}
Describe scattered sand lenses (high-permeability inclusions) in a clay matrix.\\*
- Lens orientation: $\langle$orientation$\rangle$ (aspect ratio approximately $\langle a\rangle$:1)\\
- Group centroid of the three lenses: x = $\langle x\rangle$, y = $\langle y\rangle$ (the lenses sit at three different centres)\\
- Common lens axes: $\langle$major$\rangle$ cm long and $\langle$minor$\rangle$ cm thick\\
- Three elongated lenticular bodies embedded in low-permeability clay\\[5pt]
\textbf{Layered.}
Describe a horizontally stratified subsurface permeability field.\\
- Number of sand strata: $\langle n\rangle$ distinct horizontal stratum/strata\\
- Stratum center heights from the lower boundary: $\langle c_1, \dots, c_n\rangle$ cm\\
- Corresponding stratum thicknesses: $\langle w_1, \dots, w_n\rangle$ cm\\
- Clay background containing a single high-permeability sand stratum / $\langle n\rangle$ high-permeability sand strata\\[5pt]
\textbf{Random.}
Describe a spatially correlated heterogeneous permeability field.\\
- Spatial correlation: $\langle$range-category$\rangle$ (correlation length ${\sim}\langle\sigma\rangle$ grid cells)\\
- Composition: $\langle$composition-category$\rangle$ (clay fraction = $\langle f\rangle$)\\
- Irregular clay patches distributed throughout a sandy matrix with no clear geometric structure\\[5pt]
\textbf{Continuous.}
Describe a continuously varying heterogeneous permeability field.\\
- Spatial correlation: $\langle$range-category$\rangle$ (correlation length ${\sim}\langle\sigma\rangle$ grid cells)\\
- Permeability character: $\langle$K-character$\rangle$ (mean $\log_{10}(K)$ fraction = $\langle\mu\rangle$)\\
- Smooth gradational variation in hydraulic conductivity without sharp boundaries\\[5pt]
\textbf{Shared closing instruction (all classes)}\\
Write one sentence describing this geological feature.\\[5pt]
\textbf{Additional instruction (paraphrases only)}\\
Produce an independent paraphrase with identical geological meaning and all stated attributes preserved.\\[5pt]
\textbf{Style hint (all descriptions)}\\
Style hint: $\langle$style hint$\rangle$\\[5pt]
\textbf{Shared fidelity requirement (all descriptions)}\\
Fidelity requirement: state every following value or fact explicitly and exactly in the sentence, while keeping natural prose: $\langle$required values and facts$\rangle$.
\end{promptbox}
\end{nolinenumbers}

The formatted attributes $\mathbf{a}$ in Equation~\ref{eq:dataset_generation} use fixed numerical precision: Band dip and width (fixed at $10$\,cm) are printed in integer degrees and centimetres, respectively, and centre height to $0.01$\,m; Circle mean bearing is printed in integer degrees, the sinuosity coordinate $\xi_4$ (the curvature parameter in the prompt) to $0.01$, entry height to $0.01$\,m, and segment count as an integer.
Ellipse descriptions state the orientation and the presence of three lenses, aspect ratio to $0.1$, group-centroid coordinates to $0.01$\,m, and full axis lengths to $0.1$\,cm; individual lens centres are omitted.
Layered descriptions state the integer count of active strata and their paired centre heights and thicknesses to $0.1$\,cm.
Random and Continuous correlation scales are printed to $0.1$ grid cell, and their clay fraction or pre-clipping log-mean fraction to $0.01$.
Reference values for the probes in \S\ref{sec:methods_readability} are taken from the stored attributes before prompt rounding.

The two paraphrases use different style hints from a fixed set of eight.
A rule-based validator checks sentence length, required numerical values at the stated precision, and material and geometric consistency.
Outputs that fail these checks are regenerated with feedback on rejection reasons and required values; after repeated failures, this feedback can also prescribe phrases or provide an exemplar sentence.
The additional constraints on material relationships and geometric wording, the style hints, and the complete validation and retry rules are included in the generation code distributed with the dataset \cite{darcy_dataset}.

\section{Encoding of Numeric facts}
\label{app:numeric_facts_encoding}

The Numeric facts input concatenates the class one-hot code, standardised attribute values, and binary presence indicators, in that order.
Classes and attributes follow the order in Table~\ref{tab:numeric_facts_encoding}; the raw attribute values $v_j$ use the printed precision of \ref{app:text_prompts}.
Each present attribute is standardised using the mean and population standard deviation over training descriptions in which it is present; standard deviations below $10^{-12}$ are replaced by one.
The orientation indicator remains binary and is not standardised.
The training statistics are applied unchanged to validation and test inputs; absent attributes have zero values and presence indicators.
The concatenated vector is not normalised to unit length.

\begin{table}[htbp]
  \centering
  \caption{The 26 attribute slots in the Numeric facts input, listed in input order. Units refer to values before standardisation; fractions and ratios are dimensionless.}
  \label{tab:numeric_facts_encoding}
  \begingroup\small
  \begin{tabular}{@{}llp{0.65\textwidth}@{}}
    \toprule
    Class & Slots $j$ & Attributes in order \\
    \midrule
    Band & 1--3 & Dip angle (degrees), centre height (m), fixed width (cm) \\
    Circle & 4--7 & Mean bearing (degrees), sinuosity coordinate $\xi_4$, entry height (m), segment count \\
    Ellipse & 8--13 & Orientation (0 horizontal, 1 vertical), aspect ratio, group-centroid coordinates $x$ and $y$ (m), full major and minor axis lengths (cm) \\
    Layered & 14 & Active stratum count \\
     & 15--18 & Sorted centre heights $c_1,\ldots,c_4$ (cm) \\
     & 19--22 & Corresponding thicknesses $w_1,\ldots,w_4$ (cm) \\
    Random & 23--24 & Correlation scale (grid cells), clay fraction \\
    Continuous & 25--26 & Correlation scale (grid cells), pre-clipping log-mean fraction \\
    \bottomrule
  \end{tabular}
  \endgroup
\end{table}

For Layered, the rounded centre--thickness pairs are sorted by increasing centre height, breaking ties by thickness, and their pairing is preserved when filling the slots.
For example, a stored four-stratum description gives the unstandardised Layered slots as
\[
  (v_{14},\ldots,v_{22})=(4,22.0,43.7,62.0,81.7,5.9,4.8,7.6,9.3).
\]
All nine corresponding presence indicators are one.

\bibliographystyle{elsarticle-num}
\bibliography{references}

\end{document}